%% file: PaperForReview.tex
\documentclass[10pt,twocolumn,letterpaper]{article}

\usepackage[pagenumbers]{wacv} 
\usepackage[accsupp]{axessibility}
\usepackage{graphicx}
\usepackage{amsmath}
\usepackage{amssymb}
\usepackage{booktabs}
\usepackage{times}

\usepackage{amsfonts}
\usepackage{booktabs}
\usepackage{comment}
\usepackage{color}
\usepackage{multirow}
\usepackage{array}
\usepackage[dvipsnames]{xcolor}
\usepackage{float}
\usepackage{xcolor}

\newcommand{\imagefeature}{\mathbf{X}}
\newcommand{\imagefeaturePE}{\imagefeature'}
\newcommand{\conceptmask}{\mathbf{P}} 
\newcommand{\concepttoken}{\mathbf{T}}
\newcommand{\concepttokenPE}{\concepttoken'}
\newcommand{\conceptembedding}{\mathbf{b}}

\usepackage[pagebackref,breaklinks,colorlinks]{hyperref}

\usepackage[capitalize]{cleveref}
\crefname{section}{Sec.}{Secs.}
\Crefname{section}{Section}{Sections}
\Crefname{table}{Table}{Tables}
\crefname{table}{Tab.}{Tabs.}

\def\wacvPaperID{1411} 
\def\confName{WACV}
\def\confYear{2024}

\begin{document}

\title{Framework-agnostic Semantically-aware Global Reasoning  for Segmentation}

\author{Mir Rayat Imtiaz Hossain$^{1,2}$  \qquad Leonid Sigal$^{1,2,3}$ \qquad James J. Little$^{1}$\\
$^1$University of British Columbia \qquad $^2$Vector Institute for AI \qquad $^3$Canada CIFAR AI Chair \\
\texttt{\{rayat137, lsigal, little\}@cs.ubc.ca}
}





\makeatletter
\DeclareRobustCommand\onedot{\futurelet\@let@token\@onedot}
\def\@onedot{\ifx\@let@token.\else.\null\fi\xspace}
\def\eg{\emph{e.g}\onedot} \def\Eg{\emph{E.g}\onedot}
\def\ie{\emph{i.e}\onedot} \def\Ie{\emph{I.e}\onedot}
\def\cf{\emph{cf}\onedot} \def\Cf{\emph{C.f}\onedot}
\def\etc{\emph{etc}\onedot} \def\vs{\emph{vs}\onedot}
\def\wrt{w.r.t\onedot} \def\dof{d.o.f\onedot}
\def\etal{\emph{et al}\onedot}

\def\latentregion{latent region}
\def\latentregionrep{latent token}

\makeatother

\twocolumn[{
\renewcommand\twocolumn[1][]{#1}
\vspace*{-10mm}
\maketitle

\begin{center}
    \centering
    \vspace{-0.2in}
    \includegraphics[width=0.70\textwidth]{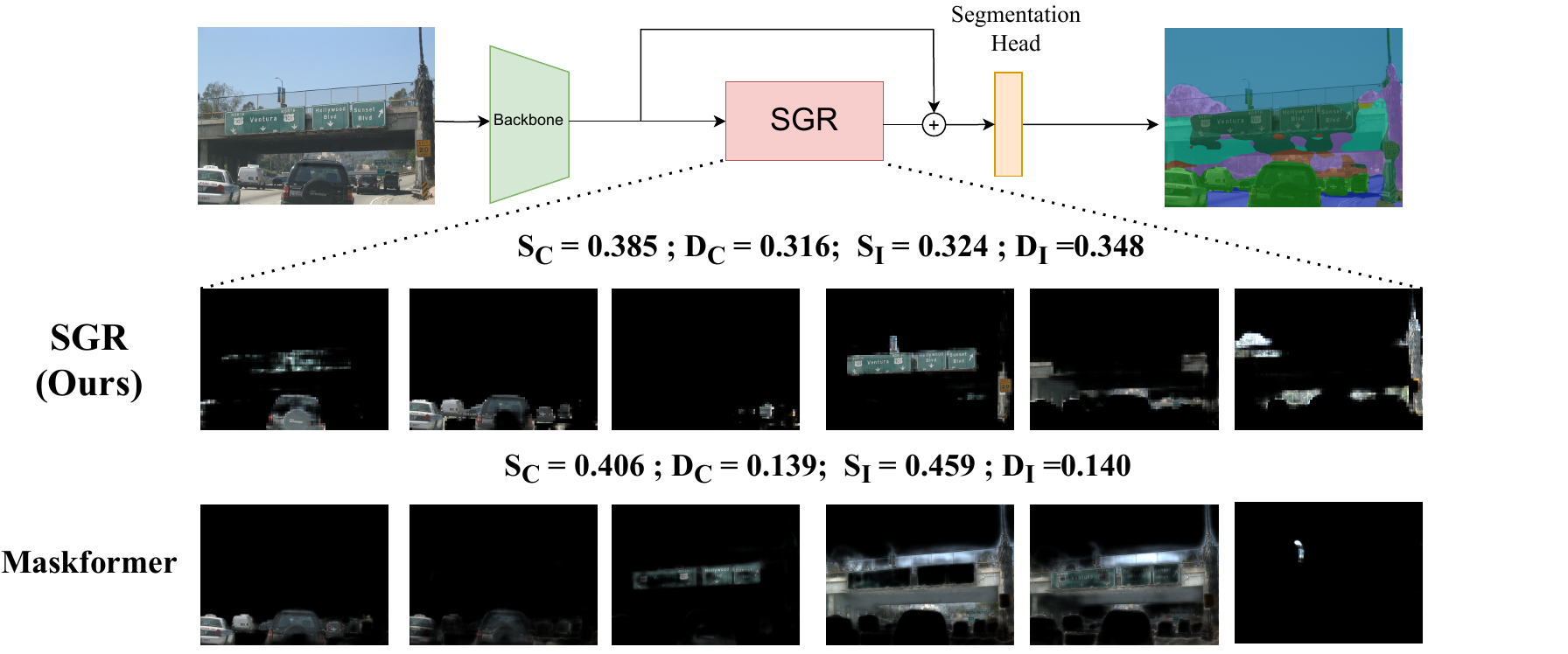}
    \vspace{-0.1in}
    \captionof{figure}{SGR  learns to group pixels into latent tokens based on semantic similarity (top); the tokens are refined using a Transformer and back-projected to enhance the original features for segmentation. Notably, the resulting tokens are more semantic than prior work, as measured by the proposed class-semantics ($S_C$) and instance-semantics ($S_I$) metrics ($\downarrow$ is better), and are more diverse ($\uparrow$ is better), as measured by proposed diversity metric at class($D_C$) and instance-level($D_I$) (see Maskformer \cite{cheng2021per} in the bottom).
    }
    \label{fig:intro-1}
\end{center}
}]

\begin{abstract}
Recent advances in pixel-level tasks (e.g. segmentation) illustrate the benefit of of long-range interactions between aggregated region-based representations that can enhance local features. However, such aggregated representations, often in the form of attention, fail to model the underlying semantics of the scene (e.g. individual objects and, by extension, their interactions). In this work, we address the issue by proposing a component that learns to project image features into latent representations and reason between them using a transformer encoder to generate contextualized and scene-consistent representations which are fused with original image features. Our design encourages the latent regions to represent semantic concepts by ensuring that the activated regions are spatially disjoint and the union of such regions corresponds to a connected object segment. The proposed semantic global reasoning (SGR) component is end-to-end trainable and can be easily added to a wide variety of backbones (CNN or transformer-based) and segmentation heads (per-pixel or mask classification) to consistently improve the segmentation results on different datasets. In addition, our latent tokens are semantically interpretable and diverse and provide a rich set of features that can be transferred to downstream tasks like object detection and segmentation, with improved performance. Furthermore, we also proposed metrics to quantify the semantics of latent tokens at both class \& instance level.
\end{abstract}

\vspace{-20pt}
\section{Introduction}
\label{sec:intro}

\begin{figure*}[t]
\centering
\includegraphics[width=\textwidth]{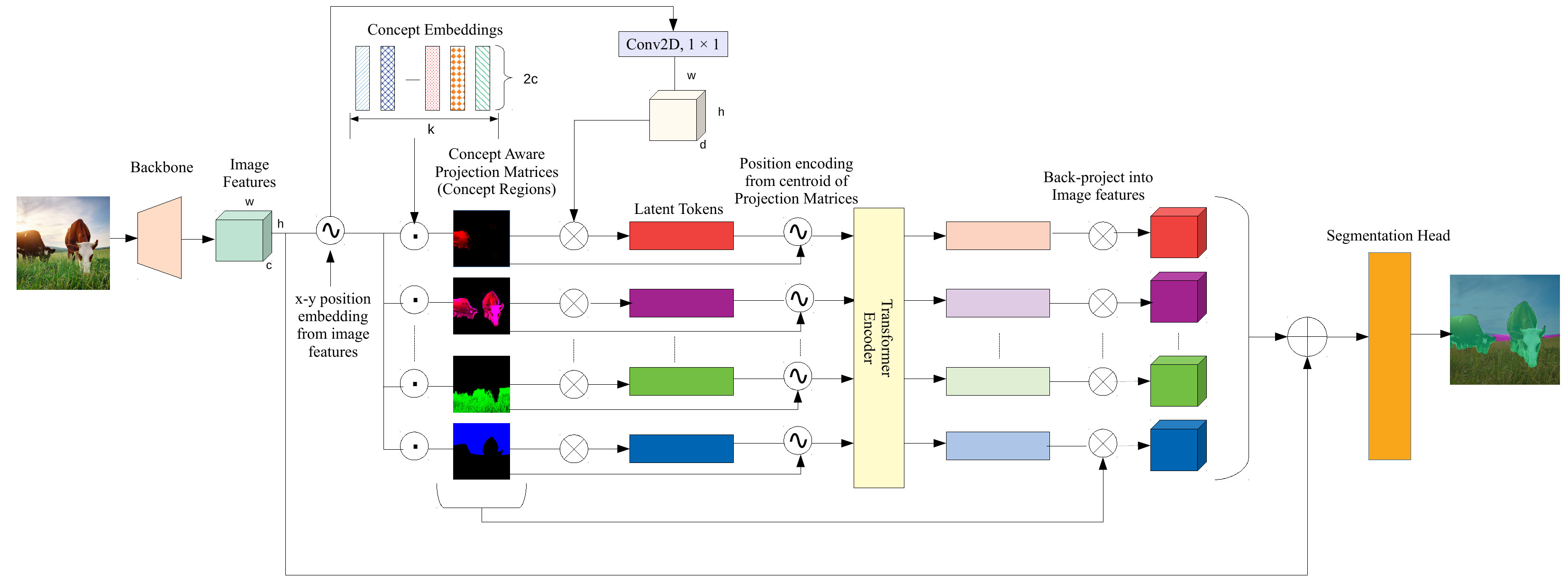} 

\caption{\textbf{Overview of our framework}. A dot-product is performed between $K$ learned concept embeddings (randomly initialized) and position-aware image features with added x-y positional embedding to generate soft projection matrices ({\em concept regions}). The projection matrices aggregate features in an object-centric manner to create latent tokens. A two-layer transformer encoder is used to reason between the tokens which are re-projected to original feature space for segmentation.}
\label{framework}
\vspace{-6pt}
\end{figure*}



Pixel-level tasks, such as semantic \cite{bertasius2017convolutional,chen2014semantic, yang2018denseaspp,zhang2019acfnet, cheng2021per, cheng2022masked, zhou2022rethinking, xie2021segformer, jain2021semask, li2022deep}, instance \cite{he2017mask, cheng2021per, cheng2021per}, and panoptic \cite{kirillov2019panoptic, cheng2022masked, cheng2021per, kirillov2023segment} segmentation, are fundamental to many computer vision problems (autonomous driving is a prime example). Recent studies~\cite{cheng2021per,chen2019graph,fu2019dual,yuan2020object} have demonstrated that even though predictions of these tasks are essentially local, incorporating global context can significantly improve performance.

Although early approaches focused on local multi-scale context where attention is computed over progressively larger patches or adding global context by pooling (\eg Parsenet~\cite{liu2015parsenet}), recent approaches have gravitated towards attentional pooling of information into latent {\em tokens}\footnote{We define {\em latent tokens} as feature representations of, not necessarily spatially contiguous, image regions.} (\eg using double-attention like in $A^2$-Net~\cite{double-attention} or ACFNet~\cite{zhang2019acfnet}) and using them to enhance and contextualize pixel representations. Other approaches like PerceiverNet~\cite{jaegle2021perceiver} or DynaPerceiver~\cite{han2023dynamic} aim at learning a dictionary of latent codes of concepts and perform a cross-attention between the latent codes and image features. The main difference between these approaches lies in how the tokens are formed, whether there are interactions between them (\eg, using graph propagation \cite{chen2019graph,li2018beyond}) and how they are aggregated back to enhance the pixel representations. Although these approaches are motivated by the overarching idea that the tokens would aggregate information over individual objects, enhancing pixels with object-centric globally-contextualized representations (\eg, through interaction between the region representations), in practice, visualizations demonstrate that the resulting latent tokens fail to capture the semantics of the scene. To address this, recent approaches have proposed to supervise the tokens directly \eg OCR~\cite{yuan2021ocnet}, Maskformer \cite{cheng2021per} and Mask2Former~\cite{cheng2022masked} use ground truth semantic mask annotations. Class-attention plays a similar role by learning class token embeddings~\cite{MCTformer}. While they result in semantically meaningful tokens and improve performance in semantic segmentation, they are limited in multiple ways. The number of tokens in these approaches \cite{cheng2021per,cheng2022masked,yuan2020object} is tied to the number of classes with each token modelling the union of all instances of a particular object in the scene. In other word, they are class-, not object-centric which can be sub-optimal since instances of an object may be different in appearance or shape or be located in different disjoint regions. Indiscernibly aggregating them may result in loss of detail. We posit that a more object-centric association (\eg where regions may be closely associated with individual instances or {\em concepts} that compose those instances) and an interaction between the tokens would provide a more granular and interpretable mechanism for attentional context. 

The core challenge for such object-centric aggregation lies in not using any instance supervision and solely relying on semantic supervision, like prior work. For the rest of the paper, we refer to the projection matrices (see Figure~\ref{framework}) which aggregate features into latent tokens as  {\em concept regions} because they aim to group features that belong to the same concept. To this end, we make following observations: \textbf{(1)} disconnected semantic segments are likely to belong to different instances -- giving us a lower bound on the number of {\em concept regions}  per image during training; \textbf{(2)} union of {\em concept regions} for a particular class should correspond to the whole semantic segmentation for the class; and \textbf{(3)} each {\em concept region} must be spatially disjoint. These constraints allow us to formulate a rich set of objectives that encourage the object-centric aggregation of features into latent tokens and refine them to add global context between the concepts, thereby enhancing the features. We show that our proposed component can be plugged into any semantic segmentation framework, regardless of the type of segmentation losses used and leads to improved performance. Additionally, the more object-centric nature of the aggregation ensures that our model learns richer features that when transferred to object detection and segmentation tasks improve performance.

\vspace{-18pt}
\paragraph{Contributions.} Our contributions are as follows:  \textbf{(i)} We propose a framework for semantically-enhanced global reasoning (SGR) that enhances local feature representations by learning to aggregate semantically similar local features into latent tokens. These tokens are then globally refined, using a Transformer Encoder~\cite{vaswani2017attention}, and re-projected into original representation. \textbf{(ii)} We propose a rich set of losses that encourage {\em concept regions} to be semantically meaningful. Specifically, we ensure that the {\em concept regions} are disjoint and the unions of them map to connected components of ground truth segments. \textbf{(iii)} We define new metrics that measure class- and instance-level semantics of the latent tokens by considering the entropy over ground truth labels that form each {\em concept region}.
\textbf{(iv)} Our component is agnostic to the backbone (CNN or transformer-based) or segmentation head and can be easily plugged into any segmentation framework. 
\textbf{(v)}   We experimentally demonstrate that adding our model to different segmentation frameworks consistently increases performance across three different benchmark datasets: Coco-Stuffs-10K, ADE-20K and Cityscapes regardless of per-pixel classification or mask-classification approach. \textbf{(vi)} Lastly, the more semantically meaningful and diverse latent representations lead to a richer feature space that improves performance when transferred to downstream tasks of object detection and segmentation.  

\section{Related Work}
\vspace{-3pt}
\paragraph{Semantic Segmentation.} Semantic segmentation has been approached by various methods in the past,  including using graph-cuts~\cite{boykov2006graph, boykov2004experimental, shi2000normalized} over specified seed points to group similar pixels and by generating mask proposals~\cite{arbelaez2012semantic, arbelaez2014multiscale,carreira2011cpmc,uijlings2013selective} and classifying them~\cite{carreira2012semantic, dai2015convolutional}. Advances in deep learning  have led to formulations of segmentation problem as a per-pixel classification~\cite{chen2014semantic, chen2017deeplab, long2015fully}. Recent studies~\cite{chen2017deeplab, chen2017rethinking, chen2018encoder, chen2019graph,li2018beyond, yuan2021ocnet, jain2021semask, xie2021segformer}  have shown that incorporating global context improves performance of semantic segmentation. More recently, approaches like Maskformer~\cite{cheng2021per} and Mask2former~\cite{cheng2021per} revived earlier mask-classification ideas, by generating mask queries and classifying them.
\vspace{-6mm}
\paragraph{Capturing Global Context.} 
Earlier approaches to capture the global context included increasing receptive fields~\cite{chen2017deeplab, chen2017rethinking, chen2018encoder} using atrous convolution~\cite{dai2017deformable, holschneider1990real,  yu2017dilated} or by aggregating features from multiple scales~\cite{chen2018encoder, yang2018denseaspp,zhao2017pyramid}.
\vspace{-6mm}
\paragraph{Global Reasoning.} Graph-based approaches and self-attention have been widely used for global reasoning between different concepts. CRFs~\cite{fields2001probabilistic, krahenbuhl2011efficient} were previously used for modeling pairwise interaction between the pixels in an image, particularly to refine predictions~\cite{bertasius2017convolutional, chandra2017dense, chen2014semantic}. Non-local Nets~\cite{wang2018non} and variants built fully connected graphs over all the pixels or a set of sampled pixels~\cite{zhang2020dynamic}  and used self-attention over the graph nodes. Recently, some approaches~\cite{chen2019graph, li2018beyond, liang2018symbolic, wu2021visual, zhang2019latentgnn} densely project the image features into latent nodes and then reason between them using graph convolutions~\cite{chen2019graph, kipf2016semi, li2018beyond} or transformers~\cite{vaswani2017attention, wu2021visual}. 
However, visualization of their latent spaces suggests failure to capture object semantics at either class or instance level. To address this, \cite{cheng2022masked, cheng2021per, yuan2020object} directly supervise latent representations. Other approaches~\cite{fu2019dual, huang2019ccnet, yuan2021ocnet} used different forms of self-attention~\cite{vaswani2017attention} to obtain the pair-wise relationship between each pixel and aggregate the information to capture the global context. Vision Transformers~\cite{dosovitskiy2020image} and  variants~\cite{caron2021emerging, liu2021swin, strudel2021segmenter, zheng2021rethinking, xie2021segformer, jain2021semask} have achieved state-of-the-art results on various computer vision tasks, including semantic segmentation, by dividing images into patches, similar to words in language models, and applying self-attention~\cite{vaswani2017attention} to reason between them. 
Subsequent approaches applied shifted windows~\cite{liu2021swin} or clustering~\cite{zeng2022not} to group the patches. Recently Dino~\cite{caron2021emerging} observed that the attention maps of class tokens tend to attend to specific regions. MCT~\cite{xu2022multi} extended~\cite{caron2021emerging} into multiple-class tokens to make them class-aware. 

\vspace{-8pt}
\section{Method}
\vspace{-6pt}
The motivation for our approach is simple: to enhance  the local features with global contextualized scene information. To this end, we introduce a semantically enhanced global reasoning (SGR) component shown in  \cref{framework}. The input to our component is a feature map from a backbone (CNN or transformer-based) with dimensions $W \times H \times C$, where $W$ and $H$ are spatial dimensions and $C$ is the channel dimension. The output is a globally contextualized and enhanced feature map of the same size.

SGR consists of four main steps. First, we compute $K$ soft masks, which we refer to as {\em concept regions}, by calculating the similarity between input features and a set of learned concept embeddings. The concept regions are weakly supervised by connected components of ground truth semantic labels. Next, the $K$ {\em concept regions} are used to project features into $K$ latent tokens, similar to context vector aggregation in soft attention. A positional embedding is added to the tokens based on the centroids of the corresponding {\em concept region}. It helps in concept disambiguation  (\eg, a low-textured blue token in the lower- and upper-part of the image can be disambiguated, permitting distinctions between {\tt sky} and {\tt water}). Third, a Transformer Encoder \cite{vaswani2017attention} is used to globally reason between the latent tokens. Finally, we re-project the refined tokens to the input feature map using the same {\em concept regions}.

\begin{figure}[t]
\centering
\includegraphics[width=\columnwidth]{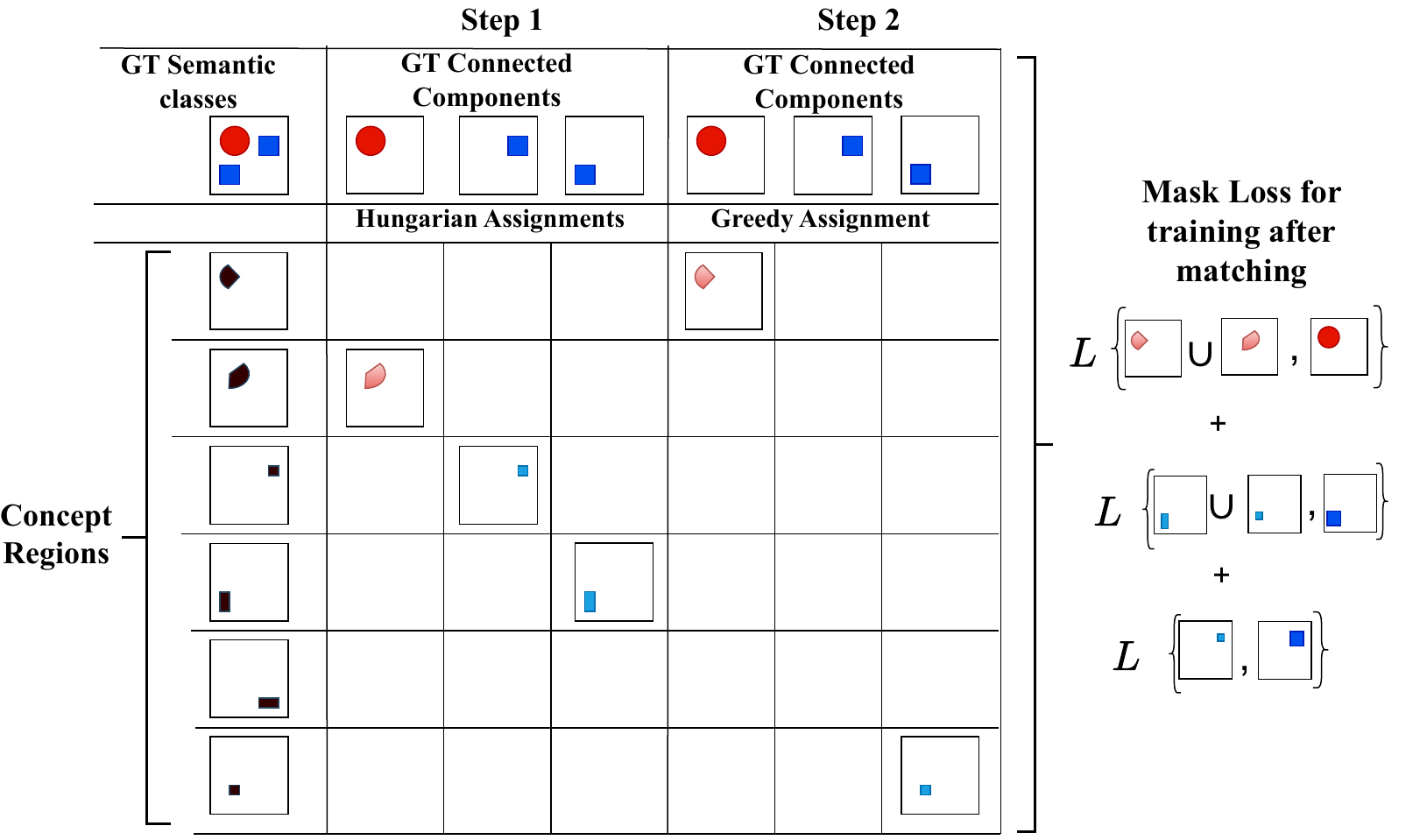} 
\caption{{\bf Visualization of our greedy matching strategy.} Binary mask losses are used to compute a cost matrix, based on which, Hungarian matching is applied to perform 1-to-1 matching between the {\em concept regions} and the ground truth components. Next, we greedily select $L$-top matches from the remaining ones. Once matched, in training, we compute losses between the union of predicted masks assigned to the same connected component. }

\label{matching}
\vspace{-6pt}
\end{figure}

The key to ensuring the ``semanticness" of our {\em concept regions}, and hence tokens, is in the first step where the $K$ concept embeddings are learned across the dataset. The concept embeddings are randomly initialized and in practice implemented by $1\times1$ convolution. Since we want our tokens to ideally represent object instances or even parts of the instances, the number of concepts we learn must be 
substantially larger than the number 
of object classes (unlike \cite{chen2019graph,cheng2021per}). Since we do not use instance-level annotations, we use connected components over semantic segmentation annotations to provide a {\em lower} bound to the number of concepts that should be active in a given training image. We assume that a subset of $L$ (of $K$) concepts can be active in each image and guide the learning of concept embeddings by matching the {\em concept regions}, that drive the latent tokens, to the class-specific connected components using a greedy matching strategy (similar to MDETR \cite{kamath2021mdetr}). Since connected components are the {\em lower} bound, the matching from {\em concept regions} to connected components is many-to-one ({\em i.e.}, multiple {\em concept regions} can be matched to a connected component). The supervision is such that the union of the {\em concept regions} that are matched to a connected component corresponds to the entire component. To ensure spatial diversity of the {\em concept regions} matched to a particular component, we minimize the pairwise cosine similarity between them. 




\subsection{Projection to Latent Semantic Tokens}
\vspace{-1pt}
The process of mapping image features into latent tokens is illustrated in Figure~\ref{framework}. Given an input feature map from a backbone, $\imagefeature \in \mathbb{R}^{W \times H \times C}$, where $W$, $H$ and $C$ are width, height and number of channels respectively,  we add positional embeddings $PE_{pos}(X)$ and $PE_{pos}(Y)$. This is done by computing sines and cosines of different frequencies for $x$- and $y$- axes of each feature cell where $PE_{pos}(x) \in \mathbb{R}^C$. The result is a positionally aware local feature tensor:
\vspace{-6pt}
\begin{equation}
\imagefeaturePE = \left[ \begin{array}{c} 
\imagefeature + PE_{pos}(X) \\
\imagefeature + PE_{pos}(Y) \\
\end{array}
\right],
\end{equation}
where $\imagefeaturePE  \in \mathbb{R}^{W \times H \times 2C}$. This allows the module to distinguish between features which are visually similar but are at different locations.  

The {\em concept regions} are obtained by computed dot product between each feature cell of  $\imagefeaturePE$ with learned concept embeddings $\conceptembedding_k \in \mathbb{R}^{2C}$;  $k \in [1, K]$ where $K$ is the total number of latent tokens. The concept embeddings are randomly initialized and are implemented as 2D convolution with $K$ kernels of size $1\times 1$. The resulting {\em concept regions} are $K$ soft masks of resolution $W \times H$, collectively forming  $\conceptmask = [\conceptmask_1, \conceptmask_2, ..., \conceptmask_K] \in \mathbb{R}^{W \times H \times K}$:

\vspace{-6pt}
\begin{equation}
\conceptmask = \text{\tt sigmoid}(\text{\tt Conv2D}(\imagefeaturePE; \{ \conceptembedding_k \}_{k=0}^K )). 
\end{equation}

Another $1 \times 1$ convolution layer is used  to reduce the dimensionality of $\imagefeaturePE$ to $\mathbb{R}^{H \times W \times D}$. The $K$ latent tokens are formed by matrix multiplication between the dimensionally reduced positional features and obtained {\em concept regions}:

\vspace{-6pt}
\begin{equation} \label{eq:1}
\concepttoken = \conceptmask^T \odot \text{\tt Conv2D}(\imagefeaturePE; \mathbf{W}_d)^T.
\end{equation}
\vspace{-6mm}

\subsection{Global Reasoning Between the Tokens}

Once we obtain latent tokens, we add positional encoding for the location of the tokens themselves into their respective representations. We characterize the position of the token by a weighted centroid computed by its concept region. 
Mainly, 
\vspace{-2pt}
$\concepttokenPE = \concepttoken + PE_{pos}(\mathbf{C})$, where
\begin{equation}
\mathbf{C}_k = \frac{1}{\sum_i \sum_j \conceptmask_{i,j,k}} \sum\limits_{w=1}^{W} \sum\limits_{h=1}^{H} \left[w \conceptmask_{w,h,k}; h \conceptmask_{w,h,k} \right].
\end{equation}
\vspace{-2pt}

The tokens with positional encoding, $\concepttokenPE$, are passed to a simple \textit{two-layer} transformer encoder\cite{vaswani2017attention} which applies self-attention over the tokens thereby performing global reasoning between them. The weights of this transformer encoder block are randomly initialized and not pre-trained. Overall,  this reasoning is analogous to identifying the relationships between different semantic concepts at different locations because each of the latent tokens correspond to a connected component of a concept. 


The output from the transformer is then re-projected onto the feature space using the same concept regions that we used to project the original features. The re-projected features are added with the input features and passed to the segmentation head.

\subsection{Concept Region Supervision}

We aim to make each token semantic, such that it only aggregates features from the component of the semantic class (ideally an instance, but possibly a part of the instance of an object)
it represents. For example, we would like to have concepts that represent each {\em car} in a scene, but would also be happy with concepts that separately represent each car's {\em wheels}, {\em body} and so forth. We observe that disconnected components of the semantic segments for a given object class are likely to belong to different instances (see supplementary) and hence they should correspond to different concepts as per the construction above. To obtain connected component ground truth for supervision, we apply connected component analysis over ground truth segmentation masks.\footnote{We apply simple morphological operations and ignore components that are smaller than 5\% of the maximum area of the connected components in the given training image.}. During training, we first match the concept region of each token to these connected components based on a cost matrix. Once matched, we supervise the concept regions using the ground truth connected components to ensure that tokens
are spatially contiguous and semantic. 



Specifically, given $K$ concept regions, we assume that only up to $L$ of them are active in any given image\footnote{Note that $K$ concepts are shared for the dataset and only a subset of those are likely to be present in any one image. We let $L$ be the maximum number of concepts that are present in an image.} and must be matched to $C$ ground truth connected components. We assume $K > L > C$ (for example in most experiments we let $K=512$ and $L=64$). We first form a $K \times C$ cost matrix and match using a two-stage procedure. First, we use the Hungarian algorithm~\cite{carion2020end,cheng2021per,kuhn1955hungarian} to perform a bipartite matching between ground truth connected components and concept regions. This ensures that each connected component is associated with at least one concept region. Second, we match the remaining $L-C$  concept regions greedily to connected components by considering the remaining $(K-C) \times C$ portion of the cost matrix. The procedure is illustrated in detail in Figure~\ref{matching}. 

Each $(i,j)$-th entry in the cost matrix measures similarity between the $i$-th {\em concept region} $\mathbf{P}_i$ and the binary ground truth mask $\mathbf{M}_j$ for $j$-th connected component.
To measure similarity we use a combination of dice loss~\cite{milletari2016diceloss} and focal loss~\cite{lin2017focal},
\begin{equation} \label{matching_cost_matrix}
Cost_{i,j} (\mathbf{M}_{j})   = \mathcal{L}_{focal} (\conceptmask_{i},\mathbf{M}_{j})
        +  \rho \mathcal{L}_{dice} (\conceptmask_{i}, \mathbf{M}_{j}). 
\end{equation}
The hyperparameter $\rho$ controls the relative weight of the two terms in the cost computation. 

Once matched, we can supervise the concept regions using the same losses used to match them. However, many-to-one matching between concept regions and ground truth connected components may lead to tokens that are duplicates to others. To avoid this, we add regularization which ensures that concept regions compete for support. This is achieved by cosine similarity loss that is applied to pairs of concept regions that are matched to the {\em same} connected component: $\mathcal{L}_{cos}(\conceptmask_{i \rightarrow j}, \conceptmask_{k \rightarrow j})$. The final loss for supervision of latent concepts can be written as follows: 
\vspace{-4pt}
\vspace{-6pt}
{\small
\begin{multline}
\label{eqn_adv_x}
        \mathcal{L}_{concept}(\conceptmask, \mathbf{M}) =  \sum\limits_{j=1}^{C}
\mathcal{L}_{focal} ( 
    \sum\limits_{i \rightarrow j} \conceptmask_{i}, \mathbf{M}_{j} ) + \\ 
    \rho
\sum\limits_{j=1}^{C}
\mathcal{L}_{dice} (
    \sum\limits_{i \rightarrow j} \conceptmask_{i}, \mathbf{M}_{j} ) + \gamma
\sum\limits_{i=1}^{L}
\sum\limits_{k=1}^{L}
\mathcal{L}_{cos}(\conceptmask_{i \rightarrow j}, \conceptmask_{k \rightarrow j}).
\vspace{-6pt}
\end{multline}
}%

Note that with a slight abuse of notation the sums in the focal and dice losses are overall concept regions $i$ that matched to one connected component $j$, and are effectively modeling the union of the concept regions matched to a given component (see Figure~\ref{matching} (right)). This, in effect, means that concepts are only weakly-supervised. The $\rho$ and $\gamma$ are the balancing parameters for the loss terms.

\vspace{-4pt}
\subsection{Final Loss}

The final loss for our model is a combination of the traditional segmentation losses like per-pixel classification or mask-classification loss depending on what approach we choose and the latent concept loss defined above:
\begin{equation}
\mathcal{L} = \mathcal{L}_{seg}(\cdot, \cdot) + \beta \mathcal{L}_{concept}(\conceptmask, \mathbf{M}),
\end{equation}
where $\mathcal{L}_{Seg}$ is a segmentation loss of choice (per-pixel or mask classification loss). Again, $\beta$ is a balancing parameter between the two terms.

\vspace{-6pt}

\input{tables/tab1}
\vspace{-2pt}
\input{tables/tab2}

\section{Metrics for interpretability} \label{sec:metric}
\vspace{-2pt}

We aim to design a component that not only improves segmentation performance but have more semantic latent representations. Hence we propose a series of metrics that measure the semantics of the tokens. Our metrics rely on two core assumptions: (1) the token is semantic if its concept region belongs to a coherent object class or instance, and (2) concept regions, on the whole, should capture as many object categories and instances as possible (\ie, be diverse).

\noindent
{\bf Semantics  ($\mathcal{S}$).} To measure semantics, we first compute a histogram for each concept region. The bins of the histogram correspond to the classes present in the image. Each  pixel of concept region votes for the label of its ground truth object class based on its projection weight. The histogram is then normalized to sum to 1. In practice, for each image, we compute $K$ discrete probability distributions that measure the empirical probability of the object class belonging to the token. We measure the entropy of these probability distributions and then average the resulting K entropies. This mean entropy is used as the measure of the semantics for the tokens for a given image. Note, the {\em lower the entropy the more semantic the token representations are} because a lower entropy indicates a uni-modal distribution suggesting that most of the token support comes from a single object class. A dataset measure can be obtained by averaging the mean entropy over all images. We call this class-semantics $\mathcal{S_C}$. We extend this metric to quantify the ability of our tokens to distinguish between individual object instances and not just classes and call it instance-semantics  $\mathcal{S_I}$. The metric can be computed in the same way except the bins of the histogram correspond to the individual instances for each of the "things" classes (a class that has instance annotations). 

\noindent
{\bf Diversity ($\mathcal{D}$)} The semantics metric defined above can be minimized by having {\em all} tokens focus on only one object class or instance. Hence it is desirable to also measure diversity of tokens. Hence, we propose the token diversity metric. To compute the metric for an image, we first calculate the mean of the normalized histograms mentioned above and the variance of the histograms. Finally we calculate the mean of the variances for all images in the dataset. The higher the variance, the more diverse the tokens are. We desire a high token diversity. Similar to semantics, diversity can also be defined at the class- ($\mathcal{D}_C$) or instance-level ($\mathcal{D}_I$). 

\vspace{-6pt}
\section{Experiments}
\vspace{-5pt}
To show the effectiveness of our framework
We conduct three sets of experiments:\textbf{ (1)} experiments where we demonstrate that adding SGR component multiple segmentation frameworks having different backbones and approaches (per-pixel or mask-classification based), leads to consistent improvement in performance across three benchmark datasets: Cityscapes~\cite{Cordts2016Cityscapes}, COCO-Stuffs-10k~\cite{caesar2018coco} and ADE-20K~\cite{zhou2017scene}; \textbf{(2)} experiments that compare semantics and diversity of our latent token representations with recent alternatives GCNET~\cite{chen2019graph} and Maskformer \cite{cheng2021per}, where we show superiority of our representations regardless of backbone or framework. Finally, \textbf{(3)} we demonstrate the effectiveness of the features learned by our component by transferring the learned weights from the semantic segmentation network and fine-tuning to downstream tasks of object detection and segmentation on the MS-COCO dataset~\cite{lin2014microsoft}. We also performed a series of ablation studies to justify our design choices.  

\vspace{-6pt}
\subsection{Datasets}
\vspace{-4pt}
\noindent
\textbf{Cityscapes}~\cite{Cordts2016Cityscapes} contains street-view images captured using a dashcam. It has 19 semantic classes labeled.

\noindent
\textbf{COCO-Stuffs-10K}~\cite{caesar2018coco} is a subset of the MS-COCO~\cite{lin2014microsoft}. It has pixel-level annotations of 171 semantic classes. 

\noindent
\textbf{ADE-20K}~\cite{zhou2017scene} is subset of the ADE20K-Full dataset containing indoor and outdoor scenes with 150 classes. 

\noindent
\textbf{MS-COCO}~\cite{lin2014microsoft} is a large-scale dataset used for object detection, segmentation, and image captioning.

\begin{table}
\vspace{1mm} 
\centering\resizebox{\columnwidth}{!}{
\begin{tabular}{@{}|p{3cm}| r |r| r| r| r | r| @{}}

\hline
Backbone & $AP_{bbox}$ & $AP_{50}$ & $AP_{75}$ & $AP_{S}$ & $AP_M$ & $AP_L$ \\
\hline
Res101-C4~\cite{he2016deep, he2017mask} & 40.29 & 59.58 & 43.37 & 22.41 & 44.94 & 55.05\\
Res101-GCNET~\cite{chen2019graph} & 38.85 & 58.82 & 42.30 & 21.62 & 42.76 & 51.88 \\
\textbf{Ours-Res101-SGR} & \textbf{41.91} & \textbf{62.79} & \textbf{45.23} & \textbf{22.95} & \textbf{46.04} & \textbf{56.32}\\
\hline
Ours (w/o token sup.)  & 33.48 & 51.72 & 36.2 & 18.73 & 36.96 & 45.84\\
\hline 
\hline

Backbone & $AP_{mask}$ & $AP_{50}$ & $AP_{75}$ & $AP_{S}$ & $AP_M$ & $AP_L$ \\
\hline
Res101-C4 & 34.88 & 56.16 & 37.27 & 15.30 & 38.32 & 53.18\\
Res101-GCNET & 34.35 & 55.08 & 36.43 & 15.07 & 38.33 & 51.37 \\
\textbf{Ours-Res101-SGR} & \textbf{37.06} & \textbf{59.28} & \textbf{39.31} & \textbf{16.30} & \textbf{40.95} & \textbf{56.19}\\
\hline
Ours (w/o token sup.)  & 29.73 & 48.75 & 31.41 & 13.55 & 33.85 & 46.41\\
\hline 
\end{tabular}
}
\caption{{\bf Transfer to Object Detection and Segmentation.} Transfer learning performance on object detection and instance segmentation using Mask-RCNN~\cite{he2017mask}. All the models trained on COCO \textbf{train2017} by us and evaluated on \textbf{val2017} for a fair comparison.}

\label{tab:object_detection}
\end{table}

\begin{table}[t]
\vspace{-2mm}
\scalebox{0.5}
\centering
\resizebox{\columnwidth}{!}{
\begin{tabular}{@{}| p{5.5cm} | r| r| r| r | r| @{}}
\hline
\multicolumn{1}{|c|}{ } & \multicolumn{3}{c|}{COCO-Stuffs-10K} & \multicolumn{2}{c|}{MS-COCO}\\

\hline
Method & mIOU & $\mathcal{S}_C$\textbf{$\downarrow$} & $\mathcal{D}_C$\textbf{$\uparrow$} & $\mathcal{S}_I$ \textbf{$\downarrow$} & $\mathcal{D}_I$ \ \textbf{$\uparrow$}  \\
\hline
Ours& \textbf{39.7} & \underline{0.226} & \textbf{0.389} & \textbf{0.315} & \underline{0.316} \\
Ours w/o  focal, dice loss and cos. sim. & 38.6  & 0.556 & 0.001 & 0.640 & 0.001\\
Ours w/o cos. sim. & 37.7 & 0.519 & 0.042 & 0.629 & 0.032
\\
Ours w graph & 39.0 & 0.246 & 0.357 & 0.345 & 0.287
\\
Ours w/o x-y emb. & 38.9 & 0.412 & 0.152 & 0.441 & 0.192\\
Ours w/o centr. emb. & 38.6 & \textbf{0.224} & 0.385 & \textbf{0.315} & \textbf{0.323}\\
Ours w/o any pos. emb. & 39.2 & 0.453 & 0.106 & 0.496 & 0.143\\
\hline
\end{tabular}
}
\vspace{-1mm}
\caption{{\bf Ablation Studies on COCO-Stuffs-10K.} All models for ablation studies use Res101 backbone trained on COCO-Stuff-10K. Instance semantics and instance diversity are reported on COCO val2017.  }
\label{tab:Ablation}
\vspace{-8pt}
\end{table}

\vspace{-4pt}    
\subsection{Implementation Details}
\vspace{-4pt}

\paragraph{Semantic Segmentation.}
For semantic segmentation, we add our SGR component after the final layer of the backbones, pre-trained on ImageNet, just before the segmentation head. For dilated FCN and DeepLabV3 heads we use the SGD optimizer with a momentum of 0.9~\cite{sutskever2013importance} and a polynomial learning rate policy where the learning rate decreases with the formula $ (1 - \frac{iter}{total\_iter})^{0.9}$ with an initial learning rate of 0.006 for Cityscapes and 0.004 for ADE-20K and Coco-Stuffs. For mask-classification approaches and Swin backbones, we used the AdamW optimizer with an initial learning rate of 1e-5. For Cityscapes, we use a batch size of 8 and a crop size of $768 \times 768$. For both Coco-Stuffs and ADE-20K we use a batch size of 16, crop size of $512 \times 512$. Each segmentation model was trained on two RTX6000 GPUs with a memory of 24 GB each. For all experiments, we multiply the initial learning rate by 10.0 for the parameters of the segmentation head and SGR component. We report both the single-scale inference and multi-scale inference with a horizontal flip at scales  0.5, 0.75, 1.0, 1.25, 1.5, and 1.75 following existing work~\cite{yuan2020object,cheng2021per,fu2019dual, chen2017rethinking}.
\vspace{-5mm}
\paragraph{Transfer to Downstream Tasks.} For transferring our model on downstream tasks, we first remove the segmentation head from the network trained on COCO-Stuffs-10K and use it as a backbone for Mask-RCNN~\cite{he2017mask} to fine-tune on MS-COCO for object detection and segmentation. We trained our model on \textbf{train2017} and evaluated on \textbf{val2017}. For a fair comparison with other backbones, we trained with the same batch size, learning rate, and iterations. We trained the models using SGD with a momentum of 0.9 and  a batch size of 8 for 270K iterations. The learning rate decreased by a factor of 0.1 at 210K and 250K iterations.

Additional implementation details are in Supplemental.

\vspace{-3pt}
\subsection{Results}
\vspace{-4pt}
\paragraph{Semantic segmentation.}
The performance of our approach on semantic segmentation is given in Table~\ref{tab:seg_result}. As observed, when our component is added on top of different segmentation frameworks, it consistently improves mIoU (mean Intersection over Union) for all three datasets for both CNN based and transformer-based backbones. It achieves an improved performance regardless of using per-pixel classification approach (like dilated-FCN or UperNet) or mask-classification approach (like Maskformer~\cite{cheng2021per} and Mask2Former~\cite{cheng2022masked}). When we add SGR with Maskformer, there is a single scale mIOU improvement of 0.9 and multi-scale mIOU imrpovement of 0.6 on COCO-Stuffs dataset. We similarly see an improvement when we add SGR to Mask2Former, which uses multi-scale features. In fact, when compared against similar backbones/frameworks, we achieve the best result on COCO-Stuffs. Even when compared to recent methods like SegFormer and ProtoSeg, which use stronger backbones (Swin-B and MiT-B4) compared to Swin-T, our approach outperforms them.

On ADE-20K, adding our component to DeepLabV3 achieves the best performance over any approaches using Resnet101 except Maskformer. For Swin-T, although the improvement on COCO-Stuffs is marginal, the improvement on ADE-20K is significant. We observe similar improvements on the Cityscapes dataset when our component is added to DeepLabV3 and Dilated-FCN; DeepLabV3 + SGR outperforms all approaches using similar backbones except Mask2Former. 

Given consistent improvements over {\em three} different baseline approaches that use different backbones, tested on {\em three} diverse datasets, we expect similar improvements to arise when SGR is added to other architectures, including variants that leverage larger backbones such as Swin-L that we were unable to train due to compute constraints. 

\begin{figure*}[t!]
\centering
\includegraphics[width=\textwidth]{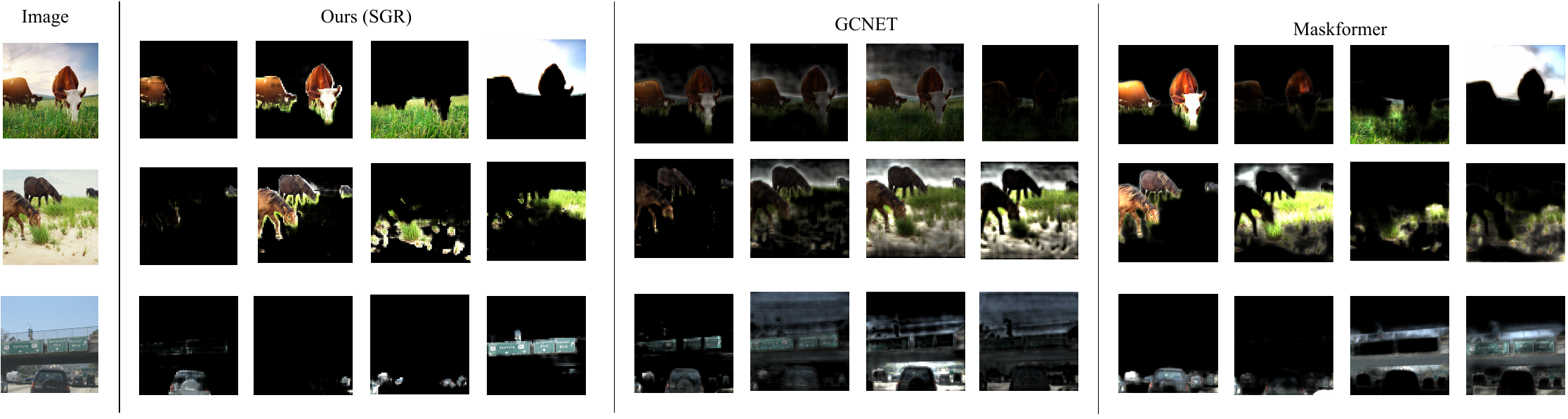} 
\caption{\textbf{Qualitative results} showing that our SGR component generates more semantically meaningful and diverse tokens. In all three images, SGR was able to disambiguate between the instances (was able to differentiate the cow in the left in the first image, and different groups/ instances of the car in the last row)  unlike Maskformer~\cite{cheng2021per}. GCNET~\cite{chen2019graph} tokens, on the other hand, lack strong semantic meaning.}
\label{qualitative}
\end{figure*}

\noindent
{\bf Class and instance-semantics.} We quantify the interpretability of the generated tokens for SGR at the class-level and instance-level using the metrics discussed in Section~\ref{sec:metric} and compare against intermediate representations of Maskformer~\cite{cheng2021per} and tokens of GLoRE~\cite{chen2019graph}. The results are shown in Table~\ref{tab:semanticness}. 
As can be observed, our tokens are more semantically meaningful than other intermediate representations at both semantic and instance levels while at the same time being more diverse. We observe this for the Swin~\cite{liu2021swin} backbone as well, showing the effectiveness of our connected component weak superivision.


\noindent
{\bf Transfer to downstream tasks.}
Table~\ref{tab:object_detection} shows the performance of our proposed component when transferred to down-stream tasks of object detection and segmentation on MS-COCO~\cite{lin2014microsoft} dataset using Mask-RCNN~\cite{he2017mask}. We compared against two other backbones, Resnet-101 (Res101-C4) pre-trained on Imagenet and GloRE (Res101-GCNET)~\cite{chen2019graph} pre-trained on COCO-Stuffs-10K on semantic segmentation task. As can be observed from Table~\ref{tab:object_detection}, SGR outperforms both on object detection and instance segmentation tasks. This demonstrates that the more semantically interpretable and diverse token representations allow us to learn richer features that are broadly more useful and transferable. 
Note that these downstream tasks require the ability to discern multiple instances and the instance-centric nature of the way our SGR aggregates information allows us to achieve this improved performance. This is further highlighted when compared against the model transferred from GloRE~\cite{chen2019graph} which also aggregates features into multiple tokens and reasons between them but the tokens lack any semantic coherence. Furthermore, as can be observed from the Table~\ref{tab:object_detection}, when we do not supervise the tokens the performance on downstream tasks drastically drops. 

\noindent
{\bf Ablation Studies.}
%
Table~\ref{tab:Ablation} highlights the importance of each of our components. As observed, when we do not supervise the tokens using mask losses, both the accuracy and semantic interpretability of tokens drop. This highlights the importance of semantically meaningful intermediate representations for better accuracy. We also observe that cosine similarity loss plays an important role in ensuring both token diversity and segmentation accuracy. Applying our losses over graph convolutions (instead of transformer encoder) allow it to have semantically meaningful tokens but the overall accuracy drops. 
Lastly we ablated importance of the two positional embeddings used. The x-y embedding is crucial to obtain object-centric tokens since it encodes spatial information. Similarly, the centroid embedding improves the accuracy. 

Ablations for hyper-parameter sensitivity (for different values of $K$ and $L$) and a discussion of computational overhead is provided in the supplementary. 

\noindent
{\bf Qualitative Results.}
Figure~\ref{qualitative} shows qualitative results for generated tokens on multiple images. As observed, our tokens are more semantically interpretable and diverse, compared to the GCNET~\cite{chen2019graph} and Maskformer~\cite{cheng2021per}. Crucially, compared to Maskformer, which also supervises tokens, SGR can distinguish between instances of objects at different spatial locations; \eg, in the first row, one of the tokens of SGR is able to distinguish the left cow from the other, while Maskformer~\cite{cheng2021per} fails to do so. We can similarly observe that SGR is able to distinguish the rightmost horse in the second image which is disjoint from the rest and in the last image three different tokens of SGR are attending to three different groups of cars where other methods failed. 

The qualitative results for semantic segmentation, object detection and segmentation are in the supplementary. 

\vspace{-3mm}
\section{Conclusion}
\vspace{-6pt}
To summarize, we propose a novel component that learns to semantically group image features into latent tokens
and reasons between them using self-attention. The losses we propose allow our tokens to distinguish between individual connected components of a semantic class. We also propose new metrics to demonstrate that our latent tokens are meaningful and semantically interpretable at both class- and instance-levels. Morever, we have empirically demonstrated that our component is independent of any backbone or segmentation framework and consistently improves performance when added with these approaches over three datasets and achieves the best performance over similar backbone and frameworks. 
Additionally, the rich set of features learned by our framework can be transferred to downstream  object detection and instance segmentation tasks. 

\footnotetext{ \textbf{Acknowledgments and Disclosure of Funding.} This work was funded, in part, by the Vector Institute for AI, Canada CIFAR AI Chair, NSERC CRC and an NSERC DG. Hardware resources used in preparing this research were provided, in part, by the Province of Ontario, the Government of Canada through CIFAR, and \href{https://vectorinstitute.ai/\#partners}{companies} sponsoring the Vector Institute. Additional support was provided by JELF CFI grant and Compute Canada under the RAC award. Finally, we sincerely thank Gaurav Bhatt for his valuable feedback on the paper draft.}

{\small
\bibliographystyle{ieee_fullname}
\bibliography{egbib}
}

\clearpage
\input{supplementary_arxiv}

\end{document}

%% file: tables/tab1.tex
\begin{table*}
\vspace{5mm}  
\centering\resizebox{1.75\columnwidth}{!}{
\begin{tabular}{@{}|p{7.0cm}| l | p{2.7cm}| r| r| @{}}
\hline
 Models & Backbone &  Datasets & mIoU (s.s.) &mIoU (m.s.) \\
  \hline
Dil-FCN$^*$ {\color{gray}[arXiv '17]} ~\cite{chen2017rethinking} & Res-101  & COCO-Stuffs & 37.8 & 38.9\\
GCNET (Dil-FCN + GloRE) {\color{gray}[CVPR '19]} $^*$~\cite{chen2019graph} & Res-101 &  COCO-Stuffs & 37.1 & 38.3\\
\textbf{Ours (Dil-FCN + SGR)} & Res-101 &  COCO-Stuffs  & \bf 38.8 (+1.0) & \bf 39.7 (+0.8) \\

Maskformer {\color{gray}[NeurIPS '21]}~\cite{cheng2021per}  &  Res-101  & COCO-Stuffs  & 38.0 &  39.3 \\
\textbf{Ours (Maskformer + SGR)} & Res-101   & COCO-Stuffs  & \bf 38.9 (+0.9) & \bf 39.9(+0.6) \\
Swin-T-Upernet$^*$ {\color{gray}[ICCV '21]} ~\cite{liu2021swin, xiao2018unified} &  Swin-T  & COCO-Stuffs  &  39.1 &  40.0 \\

\textbf{Ours (Swin-T-UperNet + SGR)} &  Swin-T  & COCO-Stuffs  & \bf 39.3 (+0.2) & \bf 40.1 (+0.1) \\

Mask2former {\color{gray}[CVPR '22]} ~\cite{cheng2022masked} &  Swin-T & COCO-Stuffs  & 42.1 &  43.0 \\

\textbf{Ours (Mask2former + SGR)} &  Swin-T  & COCO-Stuffs  & \bf 42.5 (+0.4) & \bf 43.3 (+0.3) \\

Segformer {\color{gray}[NeurIPS '21]}~\cite{xie2021segformer} &  MiT-B4 & COCO-Stuffs  & - &  42.5 \\
ProtoSeg+Segformer {\color{gray}[CVPR '22]} ~\cite{zhou2022rethinking} &  MiT-B4 & COCO-Stuffs  & - &  43.3 \\
ProtoSeg+Swin-B {\color{gray}[CVPR '22]} ~\cite{zhou2022rethinking} &  Swin-B & COCO-Stuffs  & - &  42.4 \\

\hline \hline
Dil-FCN$^*$ {\color{gray}[arXiv '17]}~\cite{chen2017rethinking} & Res-101   & ADE-20K & 42.9 & 44.0\\
GCNET  (Dil-FCN + GloRE)$^*$ {\color{gray}[CVPR '19]} ~\cite{chen2019graph} & Res-101   & ADE-20K  & 43.2  & 44.8 \\
\textbf{Ours (Dil-FCN + SGR)} & Res-101    & ADE-20K  & \bf 43.8 (+0.9) & \bf 45.6 (+1.6) \\

DeepLabV3$^*$ {\color{gray}[arXiv '17]} ~\cite{chen2017rethinking}& Res-101  & ADE-20K  &  43.1 &  44.4\\
\textbf{Ours (DeepLabV3 + SGR)} & Res-101    & ADE-20K  & \bf 44.9 (+1.8) & \bf 46.2 (+1.8) \\
Swin-T-Upernet$^*$ {\color{gray}[ICCV '21]} ~\cite{liu2021swin, xiao2018unified}  & Swin-T  & ADE-20K &  43.0 &  43.6 \\
\textbf{Ours (Swin-T-Upernet + SGR)} & Swin-T    & ADE-20K  & \bf 43.9 (+0.9) & \bf 45.0 (+1.4) \\
ProtoSeg+ HR-Net {\color{gray}[CVPR '22]}~\cite{zhou2022rethinking} &  HRNetV2-W48 & ADE-20K  & - &  43.0 \\
ProtoSeg+Swin-B {\color{gray}[CVPR '22]} ~\cite{zhou2022rethinking} &  Swin-B & ADE-20K  & - &  48.6 \\

 \hline \hline
 
Dil-FCN $^*$ {\color{gray}[arXiv '17]} ~\cite{chen2017rethinking} & Res-101   & Cityscapes-val & 77.9 & 79.3\\
GCNET (Dil-FCN + GloRE)$^*$ {\color{gray}[CVPR '19]}  ~\cite{chen2019graph} & Res-101  & Cityscapes-val & 78.0 & 79.3\\
\textbf{Ours (Dil-FCN + SGR)} & Res-101  & Cityscapes-val & \bf 78.7 (+0.7) & \bf 80.5 (+1.2) \\
DeeplabV3$^*$ {\color{gray}[arXiv '17]}~\cite{chen2017rethinking} & Res-101   & Cityscapes-val & 78.5 & 79.8\\
\textbf{Ours (DeeplabV3 + SGR)} & Res-101  & Cityscapes-val & \bf 79.8 (+1.3) & \bf 81.2 (+1.4) \\
Maskformer {\color{gray}[NeurIPS '21]}~\cite{cheng2021per} & Res-101  & Cityscapes-val & 78.5 & 80.3\\
Mask2former {\color{gray}[CVPR '22]} ~\cite{cheng2022masked} & Res-101  & Cityscapes-val & 80.1 & 81.9\\
ProtoSeg+Swin-B {\color{gray}[CVPR '22]} ~\cite{zhou2022rethinking} & Swin-B  & Cityscapes-val & - & 80.6\\

\hline
\end{tabular}
}
\caption{{\bf Results for semantic segmentation.} For Cityscapes and ADE-20K the results are reported on the validation set, while for COCO-Stuff-10K the results are reported on the test-set. The baseline models marked with $^*$ are trained by us under similar training settings as ours. The rest of the results are reported from the respective papers. Numbers within the parenthesis indicate improvement over the model on which SGR component is added. The mIoU (s.s.) indicates mIoU using single scale inference while mIoU (m.s.) indicate multi-scale inference.  }
\label{tab:seg_result}
\vspace{-6mm}
\end{table*}


%% file: tables/tab2.tex
\begin{table}
\vspace{5mm}  

\centering\resizebox{\columnwidth}{!}{

\begin{tabular}{|p{4.8cm} | l|  p{2.7cm}| r| r | }
\hline 
 Models & Backbone &  Datasets & $\mathcal{S}_C$\textbf{$\downarrow$} & $\mathcal{D}_C$\textbf{$\uparrow$} \\
\hline
GCNET  (Dil-FCN + GloRE)~\cite{chen2019graph} & Res-101 &  COCO-Stuffs  & 0.478 & 0.078\\
Maskformer~\cite{cheng2021per} & Res-101  & COCO-Stuffs  & 0.275 & 0.186  \\

\textbf{Ours (Dil-FCN + SGR)} & Res-101   & COCO-Stuffs  & \textbf{0.226} & \textbf{0.389} \\
\textbf{Ours (Swin-UpNet + SGR)} & Swin-T   & COCO-Stuffs  & 0.227 & 0.376 \\

\hline

GCNET  (Dil-FCN + GloRE)~\cite{chen2019graph} & Res-101 & ADE-20K  & 0.564 & 0.106\\



Maskformer~\cite{cheng2021per} & Res-101  & ADE-20K  & 0.335 & 0.173 \\

\textbf{Ours (Dil-FCN + SGR)} & Res-101  & ADE-20K  & 0.264 & 0.391  \\

\textbf{Ours (Swin-UpNet + SGR)} & Swin-T  & ADE-20K  & \textbf{0.240} & \textbf{0.421}  \\
 \hline
 
GCNET  (Dil-FCN + GloRE)~\cite{chen2019graph} & Res-101 &  Citys-val & 0.741 & 0.055 \\



Maskformer~\cite{cheng2021per} & Res-101 & Citys-val & 0.413 & 0.209 \\

\textbf{Ours (Dil-FCN + SGR)} & Res-101 & Citys-val & \textbf{0.293 }& \textbf{0.469} \\

\hline \hline 
 Models & Backbone  & Datasets & $\mathcal{S}_I$ \textbf{$\downarrow$} & $\mathcal{D}_I$ \ \textbf{$\uparrow$} \\
\hline
Maskformer~\cite{cheng2021per} & Res-101 & MS-COCO  & 0.383 & 0.122 \\
\textbf{Ours (Dil-FCN + SGR)} & Res-101  & MS-COCO  & \textbf{0.315} & \textbf{0.316} \\
\textbf{Ours (Swin-UpNet + SGR)} & Swin-T   & MS-COCO  & 0.328 & 0.307 \\

\hline

\end{tabular}
}

\caption{{\bf Semantics.} Comparison of class-level ($\mathcal{S}_C$) and instance-level ($\mathcal{S}_I$) semantics of intermediate tokens and the token diversity on class ($\mathcal{D}_C$) and instance-level ($\mathcal{D}_I$). The lower the value of ($\mathcal{S}_C$) and ($\mathcal{S}_I$) the more semantic the the token representations are. }
\label{tab:semanticness}
\vspace{-6pt}
\end{table}

%% file: supplementary_arxiv.tex
\setcounter{section}{0}
\setcounter{figure}{0}
\setcounter{table}{0}
\renewcommand{\thesection}{\Alph{section}}
\renewcommand{\thefigure}{\Alph{figure}}
\renewcommand{\thetable}{\alph{table}}

\twocolumn[
  \begin{center}
    \Large \textbf{Framework-agnostic Semantically-aware Global Reasoning for Segmentation} \\  
    \Large Supplementary Material
  \end{center}
]




\begin{figure}[t]
\centering
\includegraphics[width=1\columnwidth]{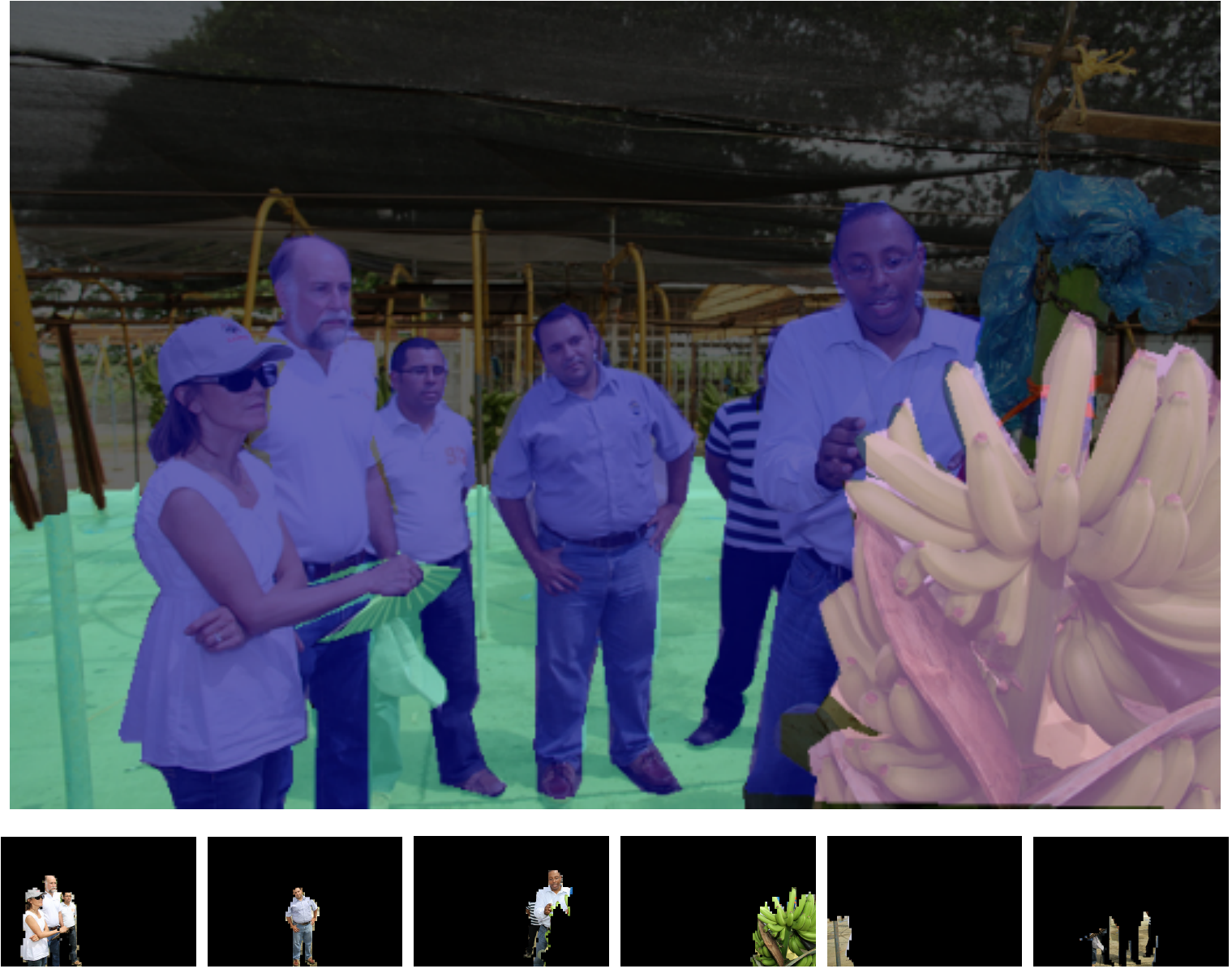} 
\caption{{\bf Ground truth segmentation mask and corresponding connected components.} The connected components are assumed to give a lower bound on the number of instances; \eg, the illustrated image contains 6 people but the number of connected components that corresponds to person is only 3.}
\label{fig:connected_comp}
\vspace{-6pt}
\end{figure}

 \section{Implementation Details}
 \iftoggle{wacvpagenumbers}{}{%
     \thispagestyle{empty}
   }
\subsection{Semantic Segmentation}

\begin{figure*}[!]
\centering
\includegraphics[width=\textwidth]{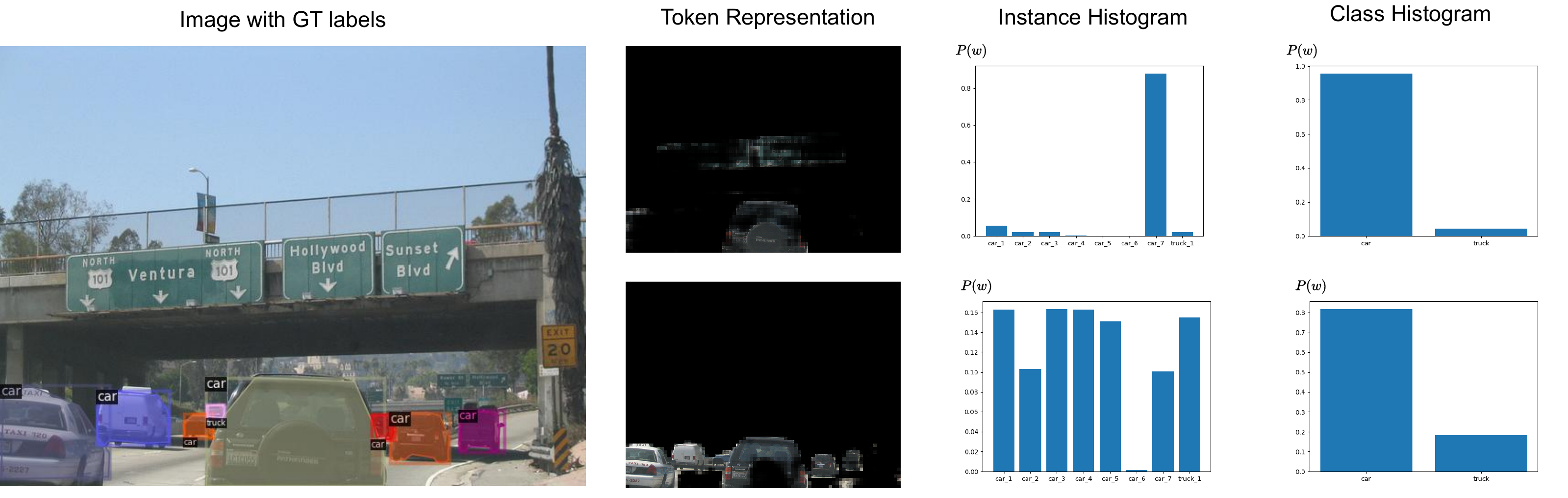} 
\caption{\textbf{Visualization of instance- and class-level histograms.} (Left) Image with ground truth instances of "things" classes. (2nd Column) Two different concept regions aggregating information. (3rd Column) Instance-level histograms. (Right) Class-level histograms. $P(w)$ indicates the probability that weight of a concept region belongs to a particular concept. While the bottom token captures a good class-level semantics, on instance-level the semantics are poor. }
\label{fig:histogram}
\end{figure*}

\begin{figure*}[!]
\vspace{-9pt}
\centering
\includegraphics[width=0.80\textwidth]{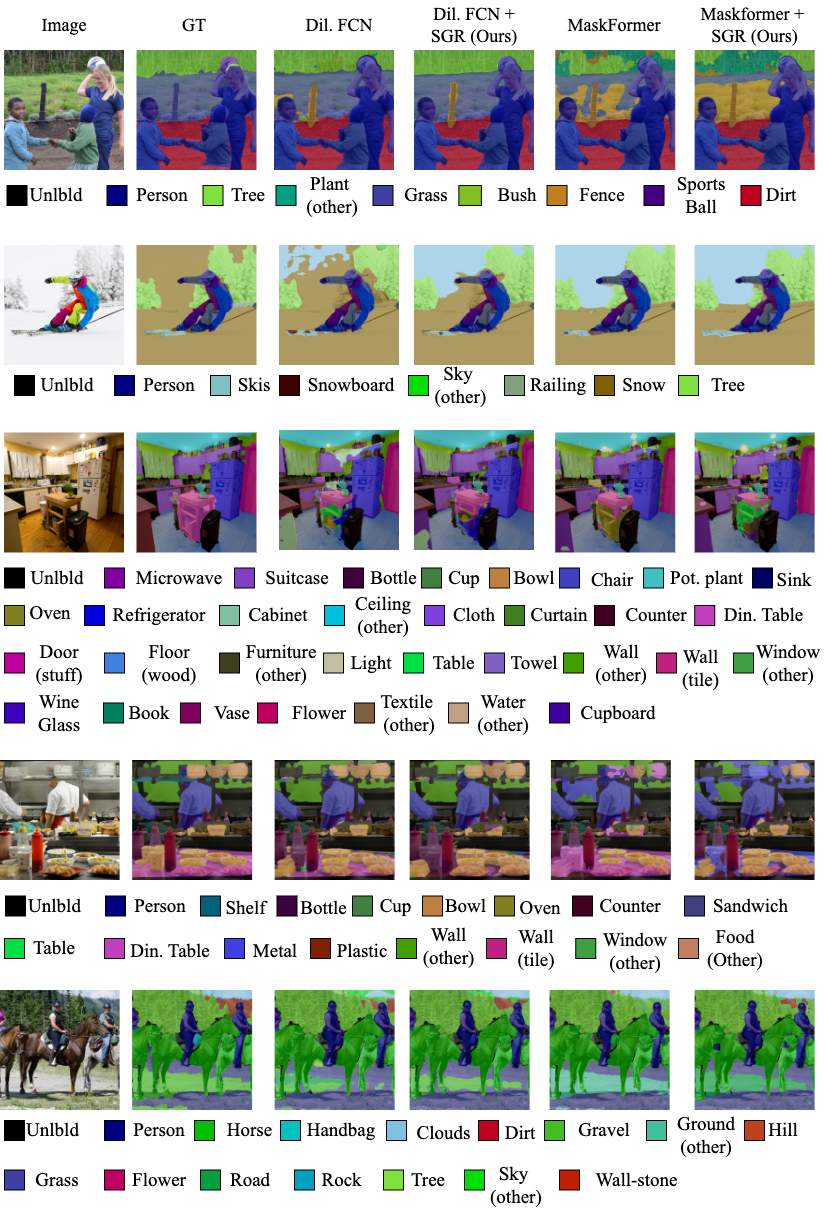} 
\caption{\textbf{Qualitative results on COCO-Stuffs-10K}. The leftmost two columns correspond to the image and ground truth semantic segmentation; the third column shows the predictions of the Dilated-FCN head; the fourth column shows the predictions of our SGR component added with Dilated-FCN~\cite{chen2017rethinking}; the fifth column shows predictions from Maskformer~\cite{cheng2021per} and the last column shows the predictions of our model added on top of Maskformer. The colors representing the class are also shown below the images.}
\label{qualitative-coco}

\end{figure*}

\begin{figure*}[!]
\centering
\includegraphics[width=\textwidth]{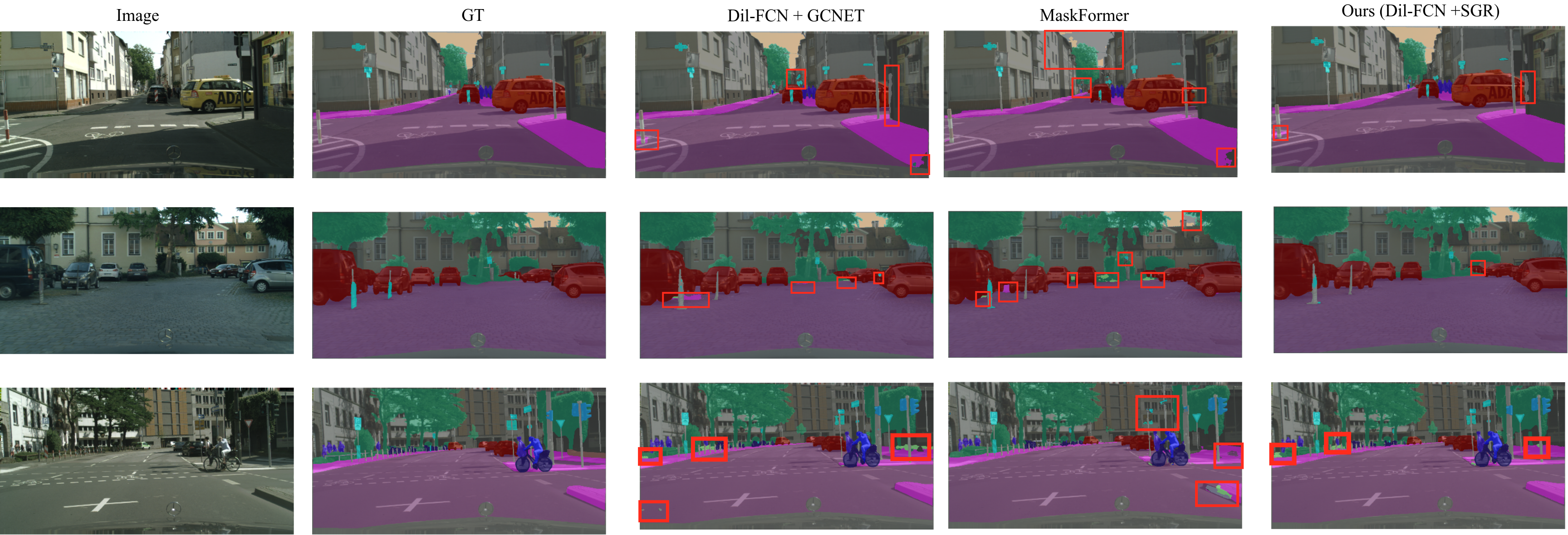} 
\caption{\textbf{Qualitative results on Cityscapes}. The leftmost two columns correspond to the image and ground truth semantic segmentation; the third column shows the predictions of Dilated-FCN+GCNET~\cite{chen2019graph}; the four column shows predictions from Maskformer~\cite{cheng2021per} and the final column shows predictions of our SGR component on top of Dilated-FCN. Red rectangles on the images indicate locations where the models failed to correctly segment the pixels.     }
\label{qualitative-city}

\end{figure*}

\begin{figure*}[!]
\centering
\includegraphics[width=\textwidth]{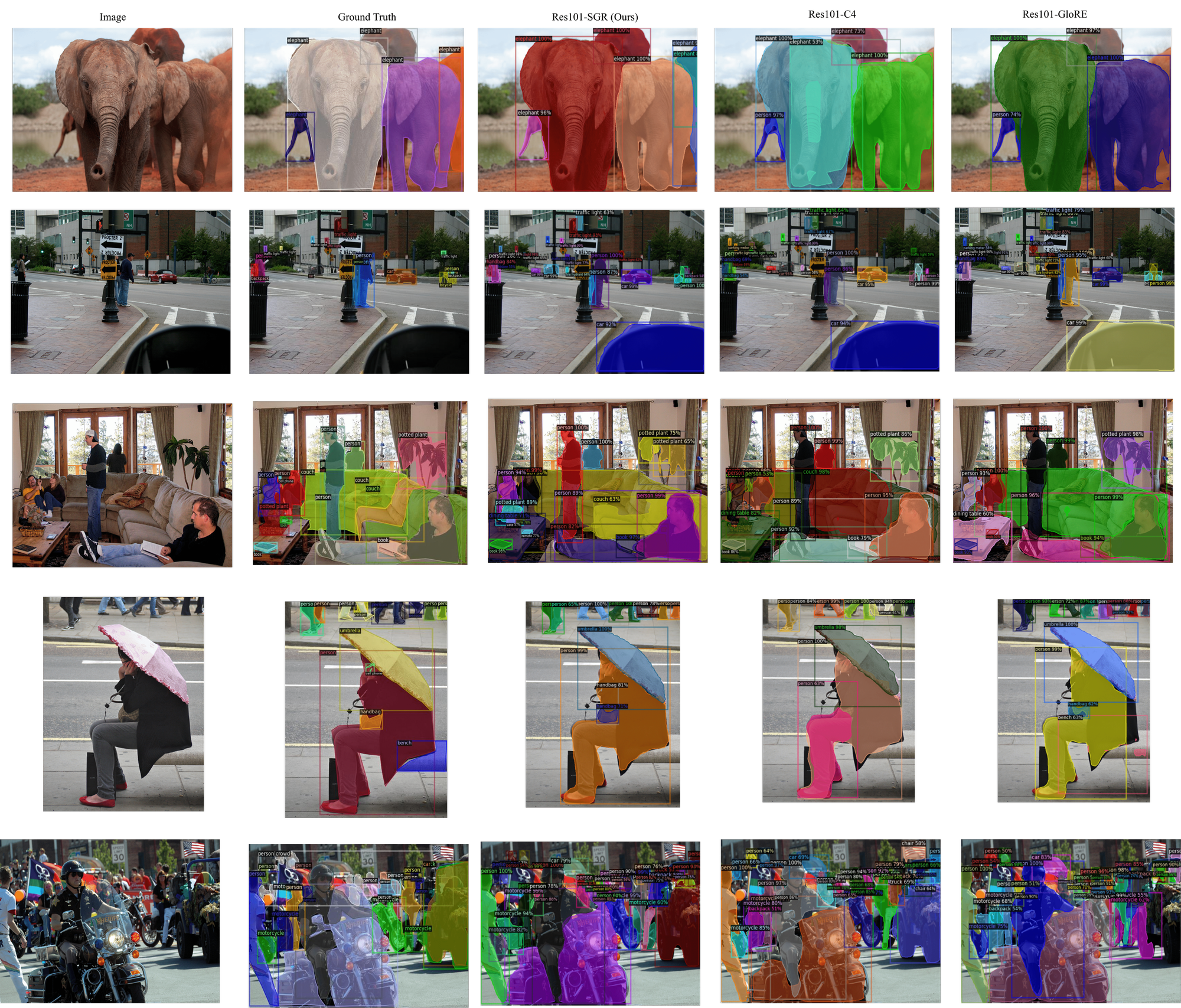} 
\caption{\textbf{Qualitative results for downstream tasks of object detection and instance segmentation on MS-COCO}. The leftmost two columns correspond to the image and ground truth object locations and their corresponding segmentation; the center column shows the predictions of using our backbone; the rightmost two columns show predictions from using Res101-C4 and Res101-GloRE~\cite{chen2019graph} backbones. }
\label{qualitative-object}

\end{figure*}

As mentioned in our main paper, for experiments involving semantic segmentation, we add our SGR component at the end of the final layer of backbones, which are pre-trained on ImageNet, before passing the processed features with added global context to appropriate segmentation heads. For the Dilated FCN~\cite{chen2017rethinking} and DeeplabV3~\cite{chen2017rethinking} heads, we use a multi-grid approach with dilated convolutions for Resnet backbones during training. The last two downsampling layers are removed, resulting in an output stride of 8. For the Swin-T~\cite{liu2021swin} backbone with UperNet~\cite{xiao2018unified}, we add the SGR component after the final Swin-T layer.

The models using Dilated-FCN and DeepLabV3 as segmentation heads are trained using the SGD optimizer with a momentum~\cite{sutskever2013importance} of 0.9 and a weight decay of 0.0001. We train on the Cityscapes dataset with an initial learning rate of 0.006 and the \textbf{ADE-20K}~\cite{zhou2017scene} and \textbf{Coco-Stuffs-10K}~\cite{caesar2018coco} datasets with an initial learning rate of 0.004. 

For the Maskformer~\cite{cheng2021per} and Mask2Former~\cite{cheng2022masked} heads, we do not use multi-grid or dilated convolution (as mentioned in the paper); hence the output features have a resolution which is 32 times smaller than the input features. The outputs of contextualized features along with rest of the layers are passed to UperNet~\cite{xiao2018unified} head for segmentation. For both Maskformer~\cite{cheng2021per} and UperNet~\cite{xiao2018unified}, we used the AdamW optimizer. For Maskformer~\cite{cheng2021per} and Mask2former~\cite{cheng2022masked} models we use the optimizers, learning rate and weight decay hyper-parameters as mentioned in the respective papers. 

For all experiments, during training, we applied random horizontal flips, random scaling between [0.5-2.0] and random color jitter following~\cite{yuan2020object,cheng2021per} for data augmentation. For \textbf{Cityscapes}~\cite{Cordts2016Cityscapes}, following the random data augmentation, the images are cropped from the center with a crop size of  $768 \times 768$. For both \textbf{ADE-20K}~\cite{zhou2017scene} and \textbf{Coco-Stuffs-10K}~\cite{caesar2018coco} a center crop of crop size $512 \times 512$ is used following the abovementioned random image transformations during training. We train models on Cityscapes using a batch size of 8 and on the other two datasets using a batch size of 16. When trained across multiple GPUs, we apply synchronized batchnorm~\cite{zhang2018context} to synchronize batch statistics following existing work~\cite{chen2017rethinking, fu2019dual,  chen2019graph, yuan2020object, cheng2021per }. We train on Cityscapes, COCO-Stuffs-10K, and ADE-20K for 240 epochs, 140 epochs and 120 epochs respectively. For all experiments, we used a polynomial learning rate policy where the learning rate decreases with the formula $ (1 - \frac{iter}{total\_iter})^{0.9}$ with every iteration.

For models with Resnet backbones the initial base learning rate is multiplied by a factor of 10.0 for the parameters of the SGR component and the layers that correspond to the segmentation head. For the Swin-T backbone we use the same learning rate for both the backbone and segmentation head.

For all three datasets, we report both the single scale inference and multi-scale inference with horizontal flip at scales  0.5, 0.75, 1.0, 1.25, 1.5 and 1.75 following existing work~\cite{yuan2020object,cheng2021per,fu2019dual, chen2017rethinking}. During multi-scale inference, the final output is calculated by taking the mean probabilities over each scale and their corresponding flipped inputs. Following ~\cite{fu2019dual, yuan2020object, cheng2021per},  for ADE-20K and COCO-Stuffs, we resize the shorter side of the image to the crop size followed by a center crop to ensure that all images are of the same size.

{\bf Hyper-parameters for training.} For Hungarian matching, in training, we used $\rho = 1.0$ for dice loss (see Eq. (5) in the main paper). For matching, as mentioned in the paper, the value of $L$ is set to 64. Hence, the top $L=64$ tokens are matched using the greedy matching approach based on the cost matrix (Figure 3 of the main paper). Once matched, we used a weight of 0.25 for the hyperparameter $\beta$ that controls the importance of binary mask losses with respect to cross-entropy loss or mask classification loss (depending on the framework we use) to train the models (see Eq. (7)).

\subsection{Transfer to Downstream Tasks}

For transfer to the downstream tasks, we removed the segmentation head from our semantic segmentation network trained on COCO-Stuffs-10K and use it as a backbone for Mask-RCNN~\cite{he2017mask} to fine-tune on the MS-COCO {\tt train2017} subset, which has 118K images, for object detection and instance segmentation. The same approach was adopted while transferring the GloRE~\cite{chen2019graph} based backbone pretrained for segmentation on COCO-Stuffs-10K. For the Res101-C4 backbone, however, we used the weights pretrained for classification on Imagenet. We reported our results on the {\tt val2017} subset having 5K images. The authors of Mask-RCNN used a batch size of 16 and trained on the {\tt trainval-135K} subset and reported results on the {\tt minival} dataset which is the same as {\tt val2017}. Therefore, for a fair comparison with other backbones, we trained them from scratch on MS-COCO {\tt train2017} using the same batch size, learning rate and iterations. We used a batch size of 8, an initial learning rate of 0.02, and used SGD with a momentum of 0.9 and weight decay of 0.0001 to train the models. We trained for 270K iterations with a learning rate decreased by 0.1 at 210K and 250K iterations. Following Mask-RCNN~\cite{he2017mask}, the RPN anchors span 5 scales and 3 aspect ratios. For all the reported backbones, 512 ROIs are sampled with a positive to negative ratio 1:3. 

\section{Ground Truth connected components} 

Figure~\ref{fig:connected_comp} shows the result of applying connected component analysis on ground truth semantic segmentation masks. As can be seen in the figure, the class {\tt person} is divided into three different components. There are altogether 6 people in total. Hence, we observe that generally connected components form a lower bound on the number of instances. Similarly, the ``stuffs" class {\tt ground} is divided into two different components and the class {\tt banana} has only one component. For ``stuffs" classes the notion of instances is not well defined, but connected components serve as a good proxy for disjoint regions that are often semantically meaningful within the scene. 

\section{Visualization of histograms for tokens}

Figure~\ref{fig:histogram} shows the visualization of class-level and instance-level histograms for two different tokens, which we use to compute class- and instance-level semantics metric (defined in the main text). The lower the entropy of each of these histograms, the more semantically meaningful the tokens are at class or instance level of granularity. As can be observed in Figure~\ref{fig:histogram}, the first token has high instance and class level semantics since it mostly aggregates information from a single car, in this case, car\_7. The lower token, despite being highly semantic at class-level (having lower entropy at class-level), is poor at capturing instance-level semantics. Hence, a token which is semantic at an instance-level is also highly semantic at class-level but not the other way around.

\begin{table}[h]
\vspace{-2mm}
\tiny 
\scalebox{0.5}
\centering
\resizebox{\columnwidth}{!}{
\begin{tabular}{@{}| l | l | r | r| r|  @{}}
\hline

\multicolumn{2}{|c|}{ Method }& FLOPs & mIOU(m.s) \\
\hline
\multirow{ 2}{*}{Dil-FCN [14]} & w/o SGR & 224G & 38.9 \\
& w SGR & 236G \textbf{(+12G)} & 39.7 \textbf{(+0.8)} \\
\hline
\multirow{ 2}{*}{MaskFormer [19] } & w/o SGR & 75G & 39.3 \\
& w SGR & 80G \textbf{(+5G)} & 39.9 \textbf{(+0.6)} \\

\hline
\end{tabular}
}

\caption{{\bf Computation Overhead for adding SGR.} Flops computed for Image size of 512 $\times$ 512. }
\label{tab:overhead}
\vspace{-8pt}
\end{table}

 \begin{table}[h]
 \tiny
\centering\resizebox{1.01\columnwidth}{!}{
\begin{tabular}{@{}| l |l| r| r| r| r| r| @{}}
\hline
K & L & mIOU (m.s.) & $S_c$ & $D_c$ & $S_I$ & $D_I$  \\
\hline
512  & 64  & 39.7 & 0.226 & 0.389 & \textbf{0.315} & \textbf{0.316} \\
512  & 32  & 39.6 & 0.242 & 0.364 & 0.344 & 0.284 \\
256 & 64 & 39.5 & \textbf{0.222} & \textbf{0.399}& 0.317 & 0.314 \\
256 & 32 & 39.7 & 0.236 & 0.376 & 0.329 & 0.297 \\
128 & 64 & \textbf{39.8} & 0.231 & 0.383 & 0.325 & 0.302 \\
\hline
 \end{tabular}}
\caption{\textbf{ Ablation for different values of K and L}. All experiments on Dil-FCN+SGR with R101 backbone}    
\label{tab:ablation_k_l}
 \end{table}

\section{Computation Overhead and Hyper-parameter sensitivity}

Table ~\ref{tab:overhead} below shows the computational overhead for our component. As can be seen, adding the SGR component consistently gives a performance boost at minimal computational burden regardless the type of framework.

We have performed an ablation for different values of K (number of tokens) and L (number of tokens matched) in Table~\ref{tab:ablation_k_l} to analyze the sensitivity of our component to those hyper-parameters. As we observe, the performance of our component is not sensitive to exact values of K and L.

\section{Qualitative Results}
\subsection{Semantic Segmentation}

\noindent
{\bf Qualitative Results on COCO-Stuffs-10K}
Figure~\ref{qualitative-coco} shows the qualitative result of semantic segmentation of on COCO-Stuffs-10K. In the first image, we observe that adding our component over Dil-FCN improves overall segmentation quality. When compared to Maskformer, our model generally misclassifies trees for plant-other, however it produces a consistent mask for fence. In fact, the ground truth is noisy in this case because there is clearly a barbed-wire fence in the image. Maskformer was able to capture the fence to a degree but produced an inconsistent map. For the second image, all methods misclassified the upper portion of the image as sky instead of snow. Compared to Dilated FCN, our component has much higher intersection over tree, person and ski classes. Maskformer makes consistent predictions however it has misclassified ski as snow-board at multiple locations. In the third image, adding the SGR component over Dilated FCN clearly produces more accurate segmentation. Maskformer misclassifies microwave and the walls and textile, which adding the SGR component improves. You can also observe that adding the SGR component also produces more consistent masks compared to Maskformer. For the fourth image, we can observe a general improvement over dilated FCN. Both Maskformer and our SGR + Maskformer perform poorly on this particular image. In the final image, Dilated FCN misclassified certain poriton of the gravel as ground-other and has generally poor intersection elsewhere (particularly for sky and hill) compared to SGR +  Dilated FCN. Maskformer has incorrectly classified the gravel as ground-other and cannot segment the hill class properly. Adding SGR over Maskformer leads to better segmentation of hill although there is still misclassification of gravel as road. 

\noindent
{\bf Qualitative Results on Cityscapes.}
Figure~\ref{qualitative-city} shows the qualitative results of semantic segmentation on Cityscapes. The red rectangles indicate locations where each of the models made incorrect predictions. In the first image, the dilated FCN misclassified the road sign and side of the pavement (rightmost corner). Both of our models (Dil-FCN +SGR and DeeplabV3+SGR) alleviate those mistakes -- DeeplabV3+SGR is more accurate. Maskformer has made incorrect classification of sky, car and pavement. We can see similar trends in the rest of the two images where our models produce more consistent and accurate predictions.

\subsection{Object Detection and Instance Segmentation}
Figure~\ref{qualitative-coco} shows the qualitative result of object detection and instance segmentation of using our pre-trained backbone in Mask-RCNN~\cite{he2017mask} on MS-COCO, compared to pre-trained Res101-C4 and Res101-GloRE~\cite{chen2019graph} backbones. In the first image, the other two backbones mis-classified the leftmost {\tt elephant} as a {\tt person} which we correctly identify and segment. Moreover, they missed the rightmost instance of the {\tt elephant} which model using our backbone was able to detect. Overall segmentation quality of each elephant was also better for our backbone. In the second image, other backbones erroneously classified the {\tt lamp} on the left as {\tt parking meter}. This is likely due to the lack of global reasoning needed to make a distinction between these two objects within the context of the scene that our backbone contains. Both of them also missed the {\tt backpack} of the person on the right. The model using our backbone consistently identifies objects and segments them better. 

In the third image, our backbone segments the {\tt couches} better than the other backbones. In the fourth image, the instance segmentation of the {\tt person} is better than the other two backbones. Moreover, the Res101-C4 backbone has missed the {\tt handbag} altogether, while the Res-101-GloRE backbone cannot segment the {\tt handbag} properly. In the final image, the Res-101-C4 backbone incorrectly labelled {\tt US flag} as a {\tt chair}. Besides, the instance segmentation quality is lower than our backbone. The Res-101-GloRE failed to identify the {\tt truck} completely, identifying part of of it as {\tt motorcycle} and inaccurately segmented it. The general quality of object segmentation is also worse. All these qualitative results demonstrate the fact that our SGR component, due to instance-like supervision through connected components, learns richer features that when transferred to downstream tasks improve performance in object detection and instance segmentation.

